\pdfoutput=1

\documentclass[11pt]{article}

\usepackage{amsmath}

\usepackage[preprint]{acl}
\usepackage{booktabs}
\usepackage{hyperref}
\definecolor{highlightgray}{gray}{0.92}
\usepackage[table,xcdraw]{xcolor} 
\definecolor{rowgray}{gray}{0.95}
\usepackage{times}
\usepackage{latexsym}
\usepackage{multirow}
\usepackage{algorithm}
\usepackage{algpseudocode} 
\usepackage{listings}

\usepackage{algpseudocode}
\usepackage{tabularx}
\usepackage{pifont}
\usepackage{tcolorbox}
\usepackage{enumitem}
\usepackage{amsmath}

\usepackage{multirow}      
\usepackage{booktabs}      
\usepackage{adjustbox}     
\usepackage{caption}       
\usepackage{geometry}      
\usepackage{amssymb}
\usepackage{graphicx}
\usepackage{subcaption}
\usepackage[T1]{fontenc}

\usepackage[utf8]{inputenc}

\usepackage{textcomp} 

\usepackage{microtype}

\usepackage{inconsolata}
\usepackage{xcolor}
\usepackage{graphicx}
\usepackage{framed}
\usepackage{pifont}
\definecolor{shadecolor}{rgb}{0.92,0.92,0.92}

\usepackage{graphicx}   
\usepackage{booktabs}   
\usepackage{wasysym}    

\PassOptionsToPackage{table}{xcolor} 
\usepackage{xcolor}
\usepackage{bbm}
\definecolor{myblue}{HTML}{418BBF}
\definecolor{myred}{HTML}{DC4748}
\definecolor{mypink}{HTML}{E78BCB}
\definecolor{mycyan}{HTML}{3BCCDB}

\newcommand{\Approach}[1]{{FineEdit-Intent}}

\title{TreeDiff: AST-Guided Code Generation with Diffusion LLMs}

\author{%
\textbf{Yiming Zeng}$^{1}$\thanks{Equal contribution.}, 
\textbf{Jinghan Cao}$^{2}$\footnotemark[1], 
\textbf{Zexin Li}$^{3}$, 
\textbf{Yiming Chen}$^{4}$,
\textbf{Tao Ren}$^{5}$, \\
\textbf{Zhuochun Li}$^{5}$, 
\textbf{Dawei Xiang}$^{1}$, 
\textbf{Xidong Wu}$^{5}$, 
\textbf{Shangqian Gao}$^{6}$, 
\textbf{Tingting Yu}$^{1}$ \\[2mm]
$^{1}$University of Connecticut,\;
$^{2}$San Francisco State University,\\
$^{3}$University of California, Riverside,\;
$^{4}$National University of Singapore,\\
$^{5}$University of Pittsburgh,\;
$^{6}$Florida State University\;
\\
}

\begin{document}
\maketitle

\begin{abstract}
Code generation is increasingly critical for real-world applications. Still, diffusion-based large language models (DLLMs) continue to struggle with this demand. Unlike free-form text, code requires syntactic precision; even minor structural inconsistencies can render a program non-executable. Existing DLLM training relies on random token masking for corruption, leading to two key failures: they lack awareness of syntactic boundaries during the iterative denoising process, and they fail to capture the long-range hierarchical dependencies essential for program correctness.
We propose TreeDiff to address both issues. Specifically, we propose a syntax-aware diffusion framework that incorporates structural priors from Abstract Syntax Tree (AST) into the corruption process. Instead of masking individual tokens at random, we selectively mask tokens belonging to key AST nodes. By aligning the corruption process with the underlying structure of code, our method encourages the model to internalize the compositional nature of programming languages, enabling it to reconstruct programs that respect grammatical boundaries and capture long-range dependencies. Our method achieves a 13.3\% relative improvement over the random masking training method, demonstrating its effectiveness in code generation by leveraging underlying structures.

\end{abstract}

\section{Introduction}

Autoregressive large language models (LLMs) have driven major advances in natural language processing and remain the prevailing approach for open-ended text generation~\cite{openai_chatgpt_blog, gemini2023technical, zeng-etal-2025-bridging, zhao2025surveylargelanguagemodels, shu2023rewrite, guo2025deepseek,chen2024three,lu2025llm,lu2024reassessing}. 
Recently, diffusion-based large language models (DLLMs) have emerged as a promising alternative to autoregressive decoding for natural language generation~\cite{li2022diffusionlm, austin2021structured, arriola2025block}. 
Instead of producing tokens strictly left-to-right, DLLMs learn to iteratively denoise corrupted sequences, enabling bidirectional context utilization and flexible conditioning~\cite{nie2025llada, austin2021d3pm, li2022diffusionlm}. 
These properties have yielded strong empirical performance on tasks such as open-domain text generation, dialogue modeling, and document completion~\cite{nie2025llada, austin2021d3pm, li2022diffusionlm}.

However, when applied to code generation, DLLMs frequently produce syntactically invalid intermediate sequences and struggle to model long-range dependencies such as variable scope and control flow~\citep{singh2023retrievaldiffusion}, leading to substantially degraded accuracy~\citep{sahoo2024mdlm, nie2025llada}.
These limitations largely stem from the random masking training paradigm, which is poorly aligned with the highly structured nature of programming languages and hinders effective generalization to complex code generation tasks.
Consequently, a significant gap remains in designing corruption and training strategies that are better suited for structured code generation.

To this end, our core insight is that \textit{for code generation, noise should not be purely stochastic}. 
Instead, incorporating prior knowledge that reflects the intrinsic structural properties of code enables DLLMs to more effectively learn and recover program-level dependencies.
Building on this insight, we introduce a syntax-aware diffusion framework that utilizes Abstract Syntax Trees (ASTs) ~\cite{AST}, hierarchical representations of source code capturing its grammatical composition, to guide masking operations during the training process. 
By aligning corruption operations with tokens associated with key AST nodes, our method encourages the model to internalize program structure, preserving local syntactic validity while modeling long-range dependencies such as scope, nesting, and control flow.

We evaluate TreeDiff across multiple code generation benchmarks, including HumanEval ~\cite{chen2021evaluatinglargelanguagemodels}, HumanEval+ ~\cite{evalplus}, MBPP ~\cite{austin2021program}, and  MBPP+ ~\cite{evalplus}, utilizing a large-scale training set of 150K samples. The results show that our syntax-aware framework consistently outperforms standard random masking strategies at various inference scales. Notably, TreeDiff achieves a \textbf{13.3\%} relative improvement on HumanEval+. Our approach also maintains stable performance during long-trajectory generation, effectively narrowing the performance gap between diffusion-based models and established autoregressive baselines.



Our contributions are as below:
\begin{itemize}
    \item To our knowledge, this is the first work to incorporate AST-aware masking into DLLMs, specifically tailored for the code generation reasoning domain. 
    \item Extensive evaluations on DLLMs demonstrate the effectiveness of our approach, which gained at most \textbf{13.3\%} relatively improvement compared with standard random masking method.
    \item  Our approach is trained on a large-scale dataset of \textbf{150K} long code reasoning samples, enabling rigorous evaluation and strong empirical support .
\end{itemize}

\section{Related Work}
\subsection{Diffusion Models for Language Modeling}
Diffusion models generate data by reversing a noise injection process, iteratively de-noising corrupted inputs \cite{ho2020denoising}. Applied to text, DLLMs reconstruct masked or noised token sequences, enabling bidirectional conditioning and flexible control compared to autoregressive decoders \cite{austin2021structured, li2022diffusionlm}. CodeFusion \cite{codefusion} applies this paradigm to code generation, iteratively denoising a complete program to overcome the ``one-way'' limitation of autoregressive models that cannot easily reconsider earlier generated tokens. 
Most existing work adopts random token-level corruption \cite{sahoo2024mdlm, nie2025llada}, which could be agnostic to rich structured information when applies to code task. To address this, CoDA~\cite{chen2025codacodinglmdiffusion} utilizes a progressive masking schedule to adapt pre-trained backbones into efficient diffusion coders. Dream-Coder~\cite{dream2025} demonstrates emergent any-order generation capability. Beyond simple distribution matching, recent work focuses on optimizing denoising trajectories to improve logical consistency. d1~\cite{zhao2025d} shows that reinforcement learning can optimize diffusion trajectories, while DiffuCoder~\cite{gong2025diffucoder} introduces a coupled-GRPO scheme to refine them via diffusion-native reinforcement learning. Both methods rely on token masking during training: d1 applies random token-level masking following the diffusion noise process, while DiffuCoder masks varying subsets of completion tokens across training passes to improve evaluation efficiency.

However, existing diffusion-based approaches still lack an explicit representation of the hierarchical structure of code. To address this, we propose an AST-guided approach that integrates program structure directly into the diffusion process.

\subsection{Abstract Syntax Tree}
An abstract syntax tree represents a program in a rooted, ordered tree data structure in which each internal node denotes a syntactic construct and each leaf typically corresponds to a terminal token \cite{aho2006compilers, parr2011antlr}. Compared to {raw source code sequences} processed linearly, AST could expose hierarchical nesting and scoping structure, providing a natural scaffold for modeling long-range dependencies in code. Previous research has exploited ASTs in several ways, such as tree-based encoders \cite{alon2019code2seq, hellendoorn2020global} and grammar-constrained decoders \cite{yin2017syntactic, rabinovich2017abstract}. Moreoever, AST-T5 \cite{AST5} employs AST subtrees as static masking templates for span corruption within T5~\cite{T5Raffel}.  

However, these approaches typically rely on autoregressive prediction or treat ASTs as static span-corruption templates, focusing on one-shot sequence completion. In contrast, we use ASTs to guide the time-dependent noise injection in diffusion, exposing structurally critical elements to masking and recovery under different noise levels during training. As a result, the diffusion trajectory is biased toward progressively reconstructing key structural components, enabling iterative refinement rather than single-step sequence completion.


\section{Method}

\subsection{Diffusion LLM}
\label{sec:method:basics}

\paragraph{Optimization Objective.}
We consider a reasoning-oriented code generation setting, where the model takes a natural language prompt $p$ as input and produces an intermediate reasoning trace $r$ together with a final executable program $c$.
We represent the full target sequence as
\[
x_0 = [\, p \,\Vert\, r \,\Vert\, c \,] \in \mathcal{V}^L,
\]
where $\mathcal{V}$ denotes a discrete token vocabulary and $L$ is the maximum sequence length.
During training, $x_0$ is sampled from a dataset $\mathcal{D}$.

Instead of autoregressive next-token prediction, we adopt a discrete diffusion formulation, where the model learns to recover $x_0$ from progressively corrupted versions.
Let $t \in \{1, \dots, T\}$ index the diffusion timestep.
The forward process samples a corrupted sequence $x_t$ via a corruption kernel
$q(x_t \mid x_0, t) = \mathrm{Corrupt}(x_0; \varepsilon_t)$,
while the denoising model $p_\theta$ is trained to reconstruct the original sequence directly from $x_t$.
Following prior work~\cite{nie2025llada}, we optimize a reconstruction objective defined only over masked positions:
\begin{multline}
\label{eq:diff_loss}
\mathcal{L}_{\text{diff}}(\theta)
=
\mathbb{E}\Bigg[
- \sum_{i=1}^{L}
\mathbbm{1}[x_t^i = \langle \texttt{mask} \rangle] \\
\cdot \log p_\theta(x_0^i \mid x_t, t)
\Bigg],
\end{multline}
where $x_0 \sim \mathcal{D}$, $t \sim \mathcal{U}(1,T)$,
and $x_t \sim q(x_t \mid x_0,t)$. This objective encourages the model to leverage the partially observed context to recover missing tokens, enabling bidirectional conditioning and iterative refinement.

\paragraph{Masking Strategy Overview.}
The corruption operator $\mathrm{Corrupt}(\cdot)$ replaces a subset of tokens in $x_0$ with a special $\langle\texttt{mask}\rangle$ symbol according to a time-dependent noise level $\varepsilon_t$.
Importantly, different semantic regions of the sequence are corrupted using different strategies.
The prompt $p$ is kept intact throughout training and serves as a fixed conditioning context.
The reasoning trace $r$, which consists of unstructured natural language, is corrupted using standard token-level random masking.
In contrast, the code region $c$ is corrupted using a structure-aware masking strategy guided by its AST, which biases the diffusion process toward preserving syntactic coherence.
We defer the detailed design of the AST-guided masking operator to Section~\ref{sec:method:ast}.

\subsection{Reasoning Chain Handling}
\label{sec:method:reasoning}
Our model explicitly incorporates intermediate reasoning steps to bridge the semantic gap between the natural language problem specification and the final executable code. We formalize the complete input sequence $x_0$ as a concatenation of three distinct regions: the prompt $p$, the reasoning chain $r$, and the target code $c$, such that $x_0 = [p \,\|\, r \,\|\, c]$.  The reasoning chain $r$ is a sequence of natural language tokens, enclosed in special tags (e.g., \texttt{<think>...</think>}), that articulates the high-level plan or logic for solving the problem described in $p$. This inclusion is inspired by chain-of-thought methodologies ~\cite{wei2022chain} to improve complex reasoning.

Given the unstructured nature of natural language, we apply a different corruption strategy to the reasoning region than the code region. During the forward diffusion process, tokens within the reasoning chain $r$ are corrupted using a standard token-level masking scheme. For a given timestep $t$, each token is independently replaced with a special $\langle\text{mask}\rangle$ token with probability $\varepsilon_t$. This approach contrasts with the structured, AST-guided span corruption applied to the code region $c$. By treating reasoning and code as distinct modalities with tailored corruption mechanisms, our model learns to denoise each region according to its unique statistical properties, capturing both the flexibility of language and the correctness of code syntax.

\begin{figure*}[t]
  \centering
\includegraphics[width=1.0\linewidth]{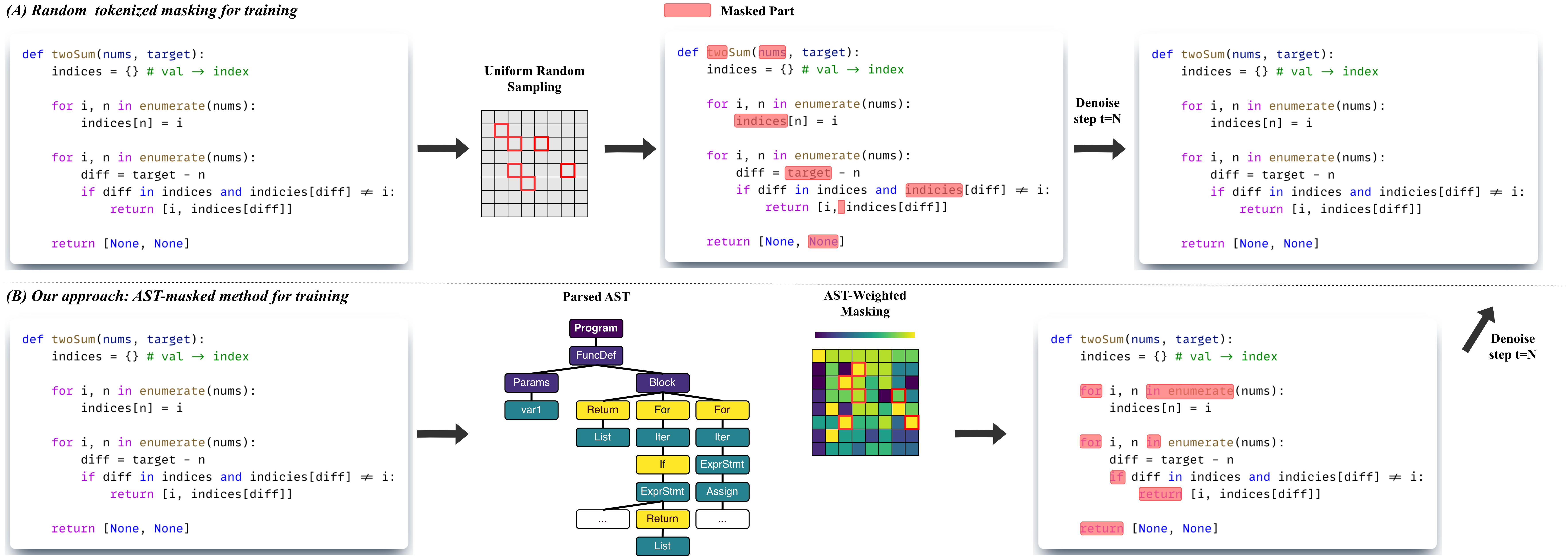}
  
  \caption{Demonstrations of the difference between our AST-Weighted Sampling Masking and Uniform Random Masking. AST-Weighted Masking is based on the AST subtrees.}
  \vspace{-2mm}
  \label{main_figure}
\end{figure*}

\subsection{ASTs as Structural Priors}
\label{sec:method:ast}

Programming languages, unlike natural language, are governed by strict syntactic rules that define their hierarchical structure. ASTs capture this structure by representing the grammatical composition of source code in tree data structures, where each node corresponds to a meaningful construct such as a statement, expression, or control block.

Formally, let $G=(V,E)$ be a tree where $V$ is the set of syntax nodes and $E$ the parent--child edges. Every node $v\in V$ has a syntactic label $\ell(v)$ drawn from a finite grammar-derived vocabulary. Leaves optionally store lexical tokens. Two auxiliary structures are often useful: (i) a linearization $\pi:V\rightarrow \{1,\dots,|V|\}$ that orders nodes for sequence-based models, and (ii) cross-tree links $R\subseteq V\times V$ encoding non-tree dependencies (e.g., data flow or symbol resolution). 
Language models can leverage $G$ to define structure-aware objectives, mask/noise schedules, and decoding constraints. 

For a more concrete example, the Python statement \texttt{x = 1} is parsed into an \texttt{ASSIGN} node with two children: a \texttt{NAME} node containing the token \texttt{"x"} and a \texttt{CONSTANT} node containing the token \texttt{"1"}. In this case, \texttt{ASSIGN}, \texttt{NAME}, and \texttt{CONSTANT} are syntactic labels, while \texttt{"x"} and \texttt{"1"} are lexical tokens. Another example is the conditional expression \texttt{if x > 0:}. Its root node is \texttt{IFSTMT}, with a subtree under the condition that includes a \texttt{COMPARE} node, which further branches into a \texttt{NAME} node with token \texttt{"x"}, a \texttt{GT} (greater-than) node, and a \texttt{CONSTANT} node with a token \texttt{"0"}.
\begin{algorithm}[t]
\footnotesize
\caption{Dynamic Tier-Aware AST Masking}
\label{alg:ast_mask}
\begin{algorithmic}[1]
\Require $x_0\in\mathcal{V}^L$ \Comment{token sequence}
\Require $\mathcal{I}_{\text{ast}}$ with Types \Comment{AST nodes and their syntactic roles}
\Require $\varepsilon_t\in[0,1]$ \Comment{target corruption rate at step $t$}
\Ensure $x_t$
\Statex
\State $N \gets \lfloor \varepsilon_t \cdot L \rfloor$ \Comment{target \#masked tokens}
\State $m \gets \mathbf{0}^L$ \Comment{init mask vector}
\Statex
\State \textbf{// Define Semantic Tiers (Weights based on text)}
\State $W \gets \mathbf{0}^L$
\ForAll{$i \in \mathcal{I}_{\text{ast}}$}
    \If{$\text{Type}(i) \in \text{Skeleton (Imports, Consts)}$} $W[i] \gets p_{\text{skel}}$
    \ElsIf{$\text{Type}(i) \in \text{DataFlow (Assigns, Calls)}$} $W[i] \gets p_{\text{data}}$
    \ElsIf{$\text{Type}(i) \in \text{CondLogic (If, Else)}$} $W[i] \gets p_{\text{cond}}$
    \ElsIf{$\text{Type}(i) \in \text{ControlFlow (Loops, Returns)}$} $W[i] \gets p_{\text{ctrl}}$
    \Else \ $W[i] \gets p_{\text{default}}$ \Comment{Default low weight for others}
    \EndIf
\EndFor
\Statex
\State \textbf{// Phase 1: Weighted Sampling based on Tiers}
\State $\mathcal{I}_{\text{cand}} \gets \text{WeightedSample}(N, \text{from}=\mathcal{I}_{\text{ast}}, \text{weights}=W)$
\ForAll{$i \in \mathcal{I}_{\text{cand}}$}
    \State $m[i] \gets 1$
\EndFor
\Statex
\State \textbf{// Phase 2: Fallback (if AST nodes are insufficient)}
\State $c \gets \text{Count}(m)$
\If{$c < N$}
    \State $\mathcal{I}_{\text{remain}} \gets \{ j \mid m[j]=0 \}$
    \State $\mathcal{I}_{\text{fill}} \gets \text{Sample}(N-c, \text{from}=\mathcal{I}_{\text{remain}})$ \Comment{Uniform random fill}
    \ForAll{$j \in \mathcal{I}_{\text{fill}}$} $m[j] \gets 1$ \EndFor
\EndIf
\Statex
\State $x_t \gets \langle\text{mask}\rangle\odot m + x_0\odot(1-m)$
\State \Return $x_t$
\end{algorithmic}
\end{algorithm}
\subsection{Dynamic AST Masking}
\label{sec:algo_desc}
Standard forward corruption in DLLMs typically applies independent Bernoulli masking at the token level. While effective for natural language, such unstructured noise risks breaking essential syntactic units, making it difficult for the model to recover the strict logical dependencies inherent in code.

To address this, our key intuition is that \textit{denoising is most effective when the noise patterns respect the data’s underlying structural logic}. We propose TreeDiff, a method that injects structural inductive biases directly into the diffusion process. As formalized in Algorithm \ref{alg:ast_mask}, TreeDiff moves beyond uniform random masking by selectively targeting syntactically critical elements. Specifically, we employ a hierarchical probability scheme derived from AST, consisting of two key phases:

\paragraph{AST Weighted Masking.}
Following the corruption rate $\varepsilon_t$, we first set a mask budget of $N = \lfloor \varepsilon_t \cdot L \rfloor$ tokens. 
As depicted in Figure \ref{main_figure}, whereas uniform random masking (top) treats all tokens as equally likely candidates for corruption, our approach (bottom) is grounded in the intuition that code tokens contribute unequally to structural integrity. 
Instead of a uniform distribution, we allocate the budget by assigning weights $W[i]$ to tokens according to their node types in the AST. 

Specifically, we define a tiered weighting scheme $\mathcal{P} = \{p_{\text{skel}}, p_{\text{data}}, p_{\text{cond}}, p_{\text{ctrl}}\}$ to prioritize different functional roles. 
We assign lower weights to structural elements ($p_{\text{skel}}$), such as imports and function definitions, to keep the high-level program structure intact. 
Conversely, we apply higher weights to logic and control flow tokens ($p_{\text{cond}}, p_{\text{ctrl}}$), such as \texttt{if} and \texttt{while} nodes, which forces the model to recover the core execution logic during the denoising process. 
Finally, the mask indices $\mathcal{I}_{\text{cand}}$ are selected via weighted sampling without replacement, ensuring that the noise distribution faithfully reflects the syntactic importance of each code element. Detailed node weights are provided in Appendix \ref{hyper_settings}.

\paragraph{Fallback Random Mechanism.} 
In scenarios where the AST parsing is incomplete, the algorithm triggers a fallback mechanism.
We calculate the current mask count $c$ and randomly sample the remaining $N - c$ tokens from the set of unmasked indices $\mathcal{I}_{\text{remain}}$. 
This hybrid design guarantees that the diffusion noise schedule is strictly adhered to, maintaining the statistical properties required for stable training while maximizing structural guidance whenever possible. Finally, the binary mask vector $m$ is applied to the input sequence $x_0$, replacing selected tokens with the special $\langle\text{mask}\rangle$ token to generate the corrupted state $x_t$.

\paragraph{Complexity Analysis.}
The computational overhead introduced by TreeDiff is negligible relative to the diffusion backbone’s forward and backward passes. As formalized in Algorithm \ref{alg:ast_mask}, our approach consists of two lightweight components: AST parsing and dynamic tier-aware weighted masking. First, given a token sequence $x_0$ of length $L$, parsing it into an AST incurs linear time complexity $O(L)$. Since $x_0$ is static across diffusion steps, this parsing can be amortized by performing it once during data preprocessing or cached efficiently for on-the-fly usage without affecting training throughput. Second, the weighted sampling strategy operates on token-level weights derived from AST node types and semantic tiers. Constructing the weight vector requires a single linear pass over AST-aligned tokens. Using efficient implementations, sampling $N$ mask indices can be performed in $O(L)$ time with linear-time weighted sampling, or $O(L \log L)$ time with heap-based methods.


\section{Evaluation}
\subsection{Experimental Setup}
\paragraph{Datasets.}
We conduct our experiments on 150,000 samples from the OpenCodeReasoning dataset~\cite{ahmad2025opencodereasoningadvancingdatadistillation}, for which negligible overlap with evaluation benchmarks (e.g., HumanEval) has been confirmed via semantic similarity checks and manual inspection. Each instance comprises three logically distinct segments: a natural language prompt, an intermediate chain-of-thought reasoning trace, and the final code solution. 


\paragraph{Evaluation Metrics. }We evaluate functional correctness using the pass@1 metric, which is defined as the proportion of tasks where the top-ranked candidate, selected by model log-likelihood, passes all unit tests. This setting ($n=1$) could reflect real-world scenarios where users prioritize the single-best prediction over multiple samples.

\begin{table*}[tbp]
\centering
\footnotesize
\setlength{\tabcolsep}{2.5pt} 
\renewcommand{\arraystretch}{1.2}

\begin{tabularx}{\textwidth}{|>{\raggedright\arraybackslash}p{3.5cm}|*{8}{>{\centering\arraybackslash}X|}}
\hline
\multirow{2}{*}{\textbf{Model}} & 
\multicolumn{4}{c|}{\boldmath{$T=256$}} & 
\multicolumn{4}{c|}{\boldmath{$T=512$}} \\ \cline{2-9} 

 & \textbf{HE} & \textbf{HE+} & \textbf{MBPP} & \textbf{MBPP+} 
 & \textbf{HE} & \textbf{HE+} & \textbf{MBPP} & \textbf{MBPP+} \\ \hline

\multicolumn{9}{|l|}{\textit{Auto-regressive Models}\textsuperscript{$\dagger$}} \\ \hline
DeepSeek-Coder-33B      & 50.6 & 44.5 & 80.4 & 70.1 & 50.6 & 44.5 & 80.4 & 70.1 \\ \hline
CodeQwen1.5-7B          & 51.8 & 45.7 & 73.5 & 60.8 & 51.8 & 45.7 & 73.5 & 60.8 \\ \hline
CodeLlama-7B            & 37.8 & 35.4 & 59.5 & 46.8 & 37.8 & 35.4 & 59.5 & 46.8 \\ \hline
CodeLlama-13B           & 42.7 & 38.4 & 63.5 & 52.6 & 42.7 & 38.4 & 63.5 & 52.6 \\ \hline

\multicolumn{9}{|l|}{\textit{Diffusion-based LLMs (LLaDA Backbone)}} \\ \hline
LLaDA-Instruct & 39.6 & 34.8 & 51.9 & 43.1 & 36.6 & 31.7 & 39.4 & 31.7 \\ \hline
+ Random Masking    & 39.6 & 35.4 & \textbf{52.9} & 43.9 & 38.4 & 32.3 & 41.5 & 32.8 \\ \hline

\multicolumn{9}{|l|}{\textit{Ours}} \\ \hline
\textbf{+ TreeDiff}                & \textbf{42.1} & \textbf{37.2} & \textbf{52.9} & \textbf{44.2} & \textbf{40.2} & \textbf{36.6} & \textbf{42.3} & \textbf{33.3} \\ \hline

\end{tabularx}
\caption{Main results on four representative benchmarks. HE and HE+ refer to HumanEval and HumanEval+, respectively. Performance is reported in Pass@1 (\%), where the denoising steps $T$ also denote the maximum generation length for each configuration. Best results are shown in bold.}
\label{tab:main_results}
\vspace{-2mm}
\end{table*}

\paragraph{Baselines.}
We evaluate the impact of different training strategies by comparing LLaDA variants with competitive baselines. All variants share the same fine-tuning data and are evaluated at two scales: 256 and 512 tokens, ensuring a comprehensive and consistent experimental setup.
\begin{itemize}[leftmargin=10px]
    \item \textbf{LLaDA-original:}  
    This baseline utilizes pretrained models, i.e.,  \texttt{LLaDA-8B-Instruct} and \texttt{LLaDA-8B-Base}, without any additional finetuning. It serves as a zero-shot reference.

    \item \textbf{LLaDA + Random Masking:}  
    We finetune the diffusion-based large language model with a standard denoising objective where tokens are uniformly masked at random across both reasoning and code regions. This setting does not incorporate any structural information and serves as a structure-agnostic baseline commonly used in language model pretraining ~\cite{nie2025llada}.

    \item \textbf{Auto-regressive Models:}  
    To provide a comparative context against standard auto-regressive architectures, we select a set of representative open-source models, including \texttt{CodeLlama} (7B/13B), \texttt{CodeQwen1.5} (7B), and \texttt{DeepSeek-Coder} (33B). These models are included to illustrate the performance landscape of widely used transformers, with results sourced directly from the EvalPlus Leaderboard~\cite{evalplus} for consistency.

\end{itemize}

\paragraph{Target Model.}
   
Our proposed method incorporates a syntax-aware masking strategy by targeting specific AST nodes instead of uniform random tokens. We apply region-specific strategies: standard random masking for the reasoning region, and AST-node-level masking for the code region. This approach provides the model with structural guidance while preserving local context to maintain stability. 


\paragraph{Implementation Details.}

The model was trained with a maximum sequence length of 4,096 tokens to balance reasoning depth with computational efficiency~\cite{yeo2025demystifying}. 
Utilizing 8 NVIDIA A100 GPUs, we employed a gradient accumulation of 16 steps to yield an effective batch size of 128. 
Based on a preliminary learning rate sweep over $\{5\times10^{-6}, 2\times10^{-5}, 5\times10^{-5}\}$. Specifically, $5\times10^{-5}$ was selected as the optimal learning rate. More training and inference details are listed in Appendix~\ref {sec:appendix_implementation}.

\paragraph{Comparison with Diffusion Baseline.}
We first evaluate the effectiveness of TreeDiff by comparing it with the \texttt{LLaDA + Random Masking} baseline, whose training method is suggested by ~\cite{nie2025llada}. As shown in Table 1, TreeDiff consistently outperforms the baseline across all metrics and denoising steps. At $T=256$, TreeDiff achieves 42.1\% / 37.2\% on HumanEval/Plus, representing a 6.3\% and 5.1\% relative improvement, respectively. It also maintains a competitive edge on MBPP/Plus, staying ahead of the random masking approach.

The performance gap becomes more pronounced at $T=512$, where the baseline suffers from significant degradation, dropping to 32.3\% on HumanEval+ and 32.8\% on MBPP+. In contrast, TreeDiff exhibits superior robustness, preserving 36.6\% on HumanEval+ and 33.3\% on MBPP+, which translates to a 13.3\% relative gain on HumanEval+. These results indicate that while random masking leads to logical instability in larger generation spaces, AST-based masking provides essential structural constraints. Notably, TreeDiff at 256 steps already surpasses the baseline's best results at 512 steps across all four metrics, confirming that structural priors significantly accelerate the convergence of diffusion-based code generation.

We further compare TreeDiff with representative open-source autoregressive (AR) models. While diffusion-based models typically lag behind the AR paradigm in code generation, TreeDiff significantly narrows this performance gap. Our method outperforms \texttt{CodeLlama-7B} (37.8\%) and achieves a performance level comparable to the larger \texttt{CodeLlama-13B} (42.7\%) on HumanEval, reaching 42.1\% at $T=256$. Although a margin remains compared to larger-scale models like \texttt{DeepSeek-Coder-33B}, these results demonstrate that incorporating structural information allows DLLMs to serve as a practical alternative for code tasks. The fact that an 8B-parameter diffusion model can match or exceed established AR baselines of similar or larger scale highlights the potency of AST-guided generation.

{
    \renewcommand{\thefootnote}{$\dagger$} 
    \footnotetext{Results are cited from the EvalPlus Leaderboard \cite{evalplus}.}
}

\subsection{Ablation Study}
\label{sec:ablation}

\noindent\textbf{Impact of Masking Granularity.} 
Table~\ref{ablation_full} compares the effectiveness of Span-level versus Token-level structural priors. 
Span-level masking corrupts contiguous code segments corresponding to complete AST subtrees rather than discrete tokens, as detailed in Appendix~\ref{alg:ast_span_v2}. 
On HumanEval, both AST-guided strategies outperform the random baseline, and TreeDiff achieves the highest pass rate of 42.1\% at $T=256$. 
However, results on MBPP reveal the limitations of coarser constraints. Span-level masking consistently underperforms the random baseline, dropping to 39.7\% compared to the baseline's 41.5\% at $T=512$. 
By contrast, TreeDiff achieves the best performance across both datasets. This indicates that while structural guidance is beneficial, the rigidity of span-based masking hinders generalization on MBPP, whereas TreeDiff offers a more robust balance between structure and flexibility.

\begin{table}[!tbp]
    \centering
    \small
    \setlength{\tabcolsep}{2pt} 
    \renewcommand{\arraystretch}{1.2}

    \begin{subtable}{1.0\columnwidth}
        \centering
        \caption{\textbf{HumanEval Results}}
        \label{tab:ablation_he}
        \begin{tabularx}{\linewidth}{|>{\raggedright\arraybackslash}p{2.4cm}|*{4}{>{\centering\arraybackslash}X|}}
        \hline
        \multirow{2}{*}{\textbf{Masking Strategy}} & \multicolumn{2}{c|}{\boldmath{$T=256$}} & \multicolumn{2}{c|}{\boldmath{$T=512$}} \\ \cline{2-5} 
         & \textbf{HE} & \textbf{HE+} & \textbf{HE} & \textbf{HE+} \\ \hline
        Random Masking & 39.6 & 35.4 & 38.4 & 32.3 \\ \hline
        AST (Span) & 40.9 & 36.6 & \textbf{40.2} & \textbf{36.6} \\ \hline
        \textbf{TreeDiff (Ours)} & \textbf{42.1} & \textbf{37.2} & \textbf{40.2} & \textbf{36.6} \\ \hline
        \end{tabularx}
    \end{subtable}

    \vspace{8pt} 

    \begin{subtable}{1.0\columnwidth}
        \centering
        \caption{\textbf{MBPP Results}}
        \label{tab:ablation_mbpp}
        \begin{tabularx}{\linewidth}{|>{\raggedright\arraybackslash}p{2.4cm}|*{4}{>{\centering\arraybackslash}X|}}
        \hline
        \multirow{2}{*}{\textbf{Masking Strategy}} & \multicolumn{2}{c|}{\boldmath{$T=256$}} & \multicolumn{2}{c|}{\boldmath{$T=512$}} \\ \cline{2-5} 
         & \textbf{MBPP} & \textbf{MBPP+} & \textbf{MBPP} & \textbf{MBPP+} \\ \hline
        Random Masking & 52.9 & 43.9 & 41.5 & 32.8 \\ \hline 
        AST (Span) & 51.6 & 42.9 & 39.7 & 31.5 \\ \hline 
        \textbf{TreeDiff (Ours)} & \textbf{52.9} & \textbf{44.9} & \textbf{42.3} & \textbf{33.3} \\ \hline 
        \end{tabularx}
    \end{subtable}

    \caption{\textbf{Ablation Study on Masking Strategies.} We compare the impact of masking granularity across different denoising steps ($T$). Table (a) reports results on HumanEval (HE) and HumanEval+ (HE+), while Table (b) reports results on MBPP and MBPP+. Best results are shown in bold.}
    \label{ablation_full}
    \vspace{-2mm}
\end{table}

\vspace{0.5em}
\noindent\textbf{Surgical vs. Coarse Constraints.} To further investigate the performance gap, we analyze the underlying mechanisms of different masking granularities. The performance drop observed with \textit{AST (Span)} masking suggests that corrupting contiguous subtrees leads to a significant loss of local information. By removing entire functional blocks, span-level masking eliminates the structural cues and dependencies required for the model to reconstruct complex logic. This is particularly evident on MBPP at $T=512$, where the span-level approach falls to 39.7\% and fails to outperform even the \textit{Random Masking} baseline of 41.5\%.

On the other hand, TreeDiff employs a more precise intervention by targeting specific syntactic nodes rather than continuous blocks. This strategy focuses on high-influence elements such as control flow nodes while keeping the surrounding context intact. As shown in the MBPP+ results, this method maintains superior stability during extended denoising. At $T=512$, TreeDiff achieves a Pass@1 of 33.3\% on MBPP+, outperforming the span-based method by 1.8\%. These results confirm that preserving the context around key nodes allows the model to maintain structural integrity while retaining the specific code patterns needed to produce executable programs.

\begin{figure}[!tbp]
\begin{minipage}{0.48\textwidth}
\begin{lstlisting}[language=Python, title={AST-Masking (Ours)}, basicstyle=\scriptsize\ttfamily, frame=single]
for char in group:
    if char == '(':
        stack.append(char)
        max_depth = max(
            max_depth, len(stack)
        )
    elif char == ')':
        if stack: stack.pop()
\end{lstlisting}
\end{minipage}
\hfill
\begin{minipage}{0.48\textwidth}
\begin{lstlisting}[language=Python, title={Random Masking}, basicstyle=\scriptsize\ttfamily, frame=single]
count = 0
for char in group:
    if char == '(':
        count += 1
    # Fails to track 
    # nested peak depth
\end{lstlisting}
\end{minipage}
\caption{Comparison of code generation results across two models on HumanEval.}
\label{fig:trajectory_compare}
\vspace{-2mm}
\end{figure}

\begin{figure*}[!tbp]
    \centering
    \begin{subfigure}{\linewidth}
        \centering
        \includegraphics[width=1.0\linewidth]{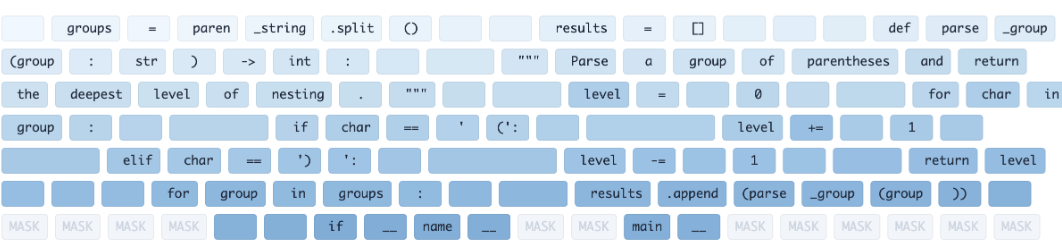}
        \caption{\textbf{Random Mask Baseline:} The model relies on a simplified counter heuristic (e.g., \texttt{level += 1}) or fragmented patterns, failing to maintain consistent syntactic anchors.}
        \label{fig:traj_sft}
    \end{subfigure}
    
    \vspace{1em} 
    
    \begin{subfigure}{\linewidth}
        \centering
        \includegraphics[width=1.0\linewidth]{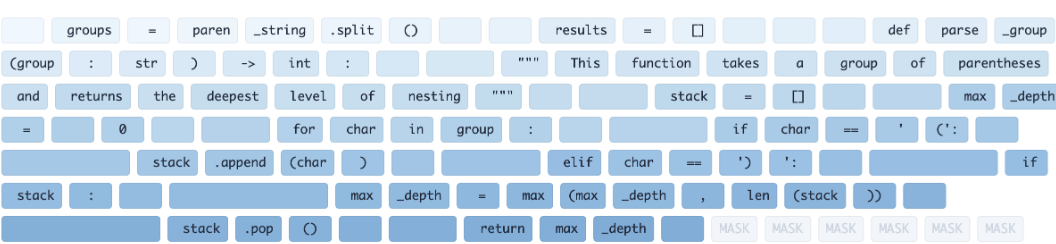}
        \caption{\textbf{TreeDiff (Ours):} The model correctly establishes a stack-based control flow (e.g., \texttt{stack.pop()}), successfully capturing the nested logic structure.}
        \label{fig:traj_ast}
    \end{subfigure}
    
    \caption{Qualitative comparison of generation trajectories at Step 110/256. The \textbf{Random Mask baseline (top)} produces fragmented code with shallow heuristics, whereas the \textbf{TreeDiff (bottom)} successfully reconstructs the complex control flow required for the task.}
    \vspace{-2mm}
    \label{fig:trajectory_comparison}
\end{figure*}
\subsection{Qualitative Study}  
The qualitative comparison of generation trajectories reveals fundamental differences in how models navigate the denoising state space when handling structured logic. As illustrated in Figure~\ref{fig:trajectory_compare}, AST-aware masking demonstrates a distinct advantage in tasks requiring deep hierarchical parsing, such as tracking the maximum depth of nested parentheses (HumanEval/6). While the random masking baseline attempts to satisfy program requirements through shallow, character-level heuristics, such as a simple counter (\texttt{count += 1}), it fails to capture the latent stateful logic required for true algorithmic correctness. This could result in fragmented code that may be syntactically plausible but lacks functional integrity in complex nesting scenarios.

In contrast, the AST-guided model exhibits structure-first emergence, correctly establishing stack-based control flow, e.g., \texttt{stack.pop()}, early in the denoising process. By selectively targeting critical syntax nodes during training, TreeDiff encourages the model to internalize the recursive nature of programming languages, treating nesting depth as a stateful property tied to tree height rather than a stochastic token-balance check. This early global scaffold provides stable constraints for subsequent refinement.

\paragraph{Generation Trajectory Analysis.}
The qualitative comparison of generation trajectories at Step 110/256 (Figure~\ref{fig:trajectory_comparison}) provides an in-depth empirical evidence for the efficacy of AST subtree sampling. We observe a fundamental divergence in how the models navigate the denoising state space: the baseline tends to recover code through local, token-level heuristics, whereas our model prioritizes the reconstruction of global structural anchors.

At this intermediate sampling stage, the baseline exhibits the limitations of uniform random masking. As shown in Figure~\ref{fig:traj_sft}, the baseline model has learned from unbiased noise. It attempts to satisfy syntax through shallow, character-level heuristics (e.g., \texttt{level += 1}); however, it fails to capture the latent nested logic of the task. This results in a fragmented trajectory where syntactic elements are generated in a disjointed, stochastic fashion. Conversely, TreeDiff demonstrates a structure-first emergence as in Figure~\ref{fig:traj_ast}. By selectively targeting high-value tokens anchored to critical AST nodes during training, the model learns to prioritize the skeleton of the program. Even at Step 110, TreeDiff has already accurately reconstructed key logical anchors such as \texttt{stack.pop()}, transforming the denoising process from a probabilistic guess into an ordered semantic refinement.

By prioritizing critical AST nodes during training, TreeDiff induces a robust anchoring effect. This mechanism effectively optimizes the generation trajectory of syntactic elements, ensuring that core topological nodes like \texttt{stack} operations and control flow manifest earlier. Such early establishment of a global scaffold provides stable constraints for subsequent denoising.

.

\section{Conclusion}
We present TreeDiff, a syntax-aware diffusion framework that addresses the limitations of random token masking in code generation by incorporating AST structural priors. By selectively masking tokens tied to key syntactic nodes, our method encourages the model to internalize the hierarchical and compositional nature of programming languages. Experimental results demonstrate that TreeDiff achieves a 13.3\% relative improvement over standard random masking methods.  Qualitative analysis confirms that TreeDiff effectively preserves local syntactic coherence while capturing critical long-range dependencies, enabling the reconstruction of complex logic that structure-agnostic models fail to resolve. 
\clearpage

\section*{Limitations}
\noindent \textbf{Inference Reasoning Chain Length.} The significant inference latency inherent to LlaDa’s iterative refinement process poses scalability challenges for extremely long sequences. Consequently, we maintain a maximum trace length of 1,024 tokens to balance experimental throughput with qualitative depth.

\noindent \paragraph{Training Tradeoff.} Full-parameter fine-tuning of LLaDA 8B model on long reasoning traces (4096) exceeds the memory capacity of our 8 NVIDIA A100 GPUs. To overcome this hardware bottleneck and ensure computational feasibility, we employ LoRA for parameter-efficient adaptation.



\bibliography{anthology,custom}

@article{lu2025llm,
  title={From LLM-anation to LLM-orchestrator: Coordinating Small Models for Data Labeling},
  author={Lu, Yao and Ji, Zhaiyuan and Du, Jiawei and Shanqing, Yu and Xuan, Qi and Zhou, Tianyi},
  journal={arXiv preprint arXiv:2506.16393},
  year={2025}
}

@article{lu2024reassessing,
  title={Reassessing layer pruning in llms: New insights and methods},
  author={Lu, Yao and Cheng, Hao and Fang, Yujie and Wang, Zeyu and Wei, Jiaheng and Xu, Dongwei and Xuan, Qi and Yang, Xiaoniu and Zhu, Zhaowei},
  journal={arXiv preprint arXiv:2411.15558},
  year={2024}
}

@inproceedings{chen2024three,
  title={A Three-Phases-LORA Finetuned Hybrid LLM Integrated with Strong Prior Module in the Education Context},
  author={Chen, Zhangquan and Liu, Chunjiang and Duan, Haobin},
  booktitle={International Conference on Artificial Neural Networks},
  pages={235--250},
  year={2024},
  organization={Springer}
}

@misc{openai_chatgpt_blog,
  title        = {Introducing ChatGPT},
  author       = {{OpenAI}},
  howpublished = {\url{https://openai.com/blog/chatgpt}},
  year         = {2022},
  note         = {Accessed: 2025-07-24}
}

@misc{gemini2023technical,
  title        = {Gemini: A Family of Highly Capable Multimodal Models},
  author       = {Team, Gemini and Anil, Rohan and others},
  howpublished = {arXiv preprint arXiv:2312.11805},
  year         = {2023}
}

@inproceedings{AST,
author = {Neamtiu, Iulian and Foster, Jeffrey S. and Hicks, Michael},
title = {Understanding source code evolution using abstract syntax tree matching},
year = {2005},
isbn = {1595931236},
publisher = {Association for Computing Machinery},
address = {New York, NY, USA},
url = {https://doi.org/10.1145/1083142.1083143},
doi = {10.1145/1083142.1083143},
booktitle = {Proceedings of the 2005 International Workshop on Mining Software Repositories},
pages = {1–5},
numpages = {5},
location = {St. Louis, Missouri},
series = {MSR '05}
}

@inproceedings{austin2021d3pm,
  title     = {Structured Denoising Diffusion Models in Discrete State-Spaces},
  author    = {Austin, Jacob and Johnson, Daniel D. and Ho, Jonathan and Tarlow, Daniel and van den Berg, Rianne},
  booktitle = {Advances in Neural Information Processing Systems},
  year      = {2021},
  url       = {https://arxiv.org/abs/2107.03006}
}

@inproceedings{li2022diffusionlm,
  title     = {Diffusion-LM Improves Controllable Text Generation},
  author    = {Li, Xiang Lisa and Thickstun, John and Gulrajani, Ishaan and Liang, Percy and Hashimoto, Tatsunori B.},
  booktitle = {Advances in Neural Information Processing Systems},
  year      = {2022},
  url       = {https://arxiv.org/abs/2205.14217}
}

@inproceedings{ho2020denoising,
  title     = {Denoising Diffusion Probabilistic Models},
  author    = {Ho, Jonathan and Jain, Ajay and Abbeel, Pieter},
  booktitle = {Advances in Neural Information Processing Systems},
  year      = {2020},
  eprint    = {2006.11239},
  archivePrefix = {arXiv}
}

@article{yeo2025demystifying,
  title={Demystifying long chain-of-thought reasoning in llms},
  author={Yeo, Edward and Tong, Yuxuan and Niu, Morry and Neubig, Graham and Yue, Xiang},
  journal={arXiv preprint arXiv:2502.03373},
  year={2025}
}

@inproceedings{alon2019code2seq,
  title     = {code2seq: Generating Sequences from Structured Representations of Code},
  author    = {Alon, Uri and Brody, Shaked and Levy, Omer and Yahav, Eran},
  booktitle = {International Conference on Learning Representations},
  year      = {2019},
  note      = {arXiv:1808.01400}
}

@inproceedings{hellendoorn2020global,
  title     = {Global Relational Models of Source Code},
  author    = {Hellendoorn, Vincent~J. and Maniatis, Petros and Singh, Rishabh and Sutton, Charles and Bieber, David},
  booktitle = {International Conference on Learning Representations},
  year      = {2020}
}

@inproceedings{yin2017syntactic,
  title     = {A Syntactic Neural Model for General-Purpose Code Generation},
  author    = {Yin, Pengcheng and Neubig, Graham},
  booktitle = {Proceedings of the 55th Annual Meeting of the Association for Computational Linguistics},
  year      = {2017},
  note      = {arXiv:1704.01696}
}

@inproceedings{rabinovich2017abstract,
  title     = {Abstract Syntax Networks for Code Generation and Semantic Parsing},
  author    = {Rabinovich, Maxim and Stern, Mitchell and Klein, Dan},
  booktitle = {Proceedings of the 55th Annual Meeting of the Association for Computational Linguistics},
  year      = {2017},
  note      = {arXiv:1704.07535}
}

@misc{ahmad2025opencodereasoningadvancingdatadistillation,
      title={OpenCodeReasoning: Advancing Data Distillation for Competitive Coding}, 
      author={Wasi Uddin Ahmad and Sean Narenthiran and Somshubra Majumdar and Aleksander Ficek and Siddhartha Jain and Jocelyn Huang and Vahid Noroozi and Boris Ginsburg},
      year={2025},
      eprint={2504.01943},
      archivePrefix={arXiv},
      primaryClass={cs.CL},
      url={https://arxiv.org/abs/2504.01943}, 
}

@book{aho2006compilers,
  title     = {Compilers: Principles, Techniques, and Tools (2nd Edition)},
  author    = {Aho, Alfred~V. and Lam, Monica~S. and Sethi, Ravi and Ullman, Jeffrey~D.},
  publisher = {Addison-Wesley},
  year      = {2006}
}

@book{parr2011antlr,
  title     = {The Definitive ANTLR 4 Reference},
  author    = {Parr, Terence},
  publisher = {Pragmatic Bookshelf},
  year      = {2011}
}

@inproceedings{austin2021structured,
  title={Structured denoising diffusion models in discrete state-spaces},
  author={Austin, Jacob and Johnson, Daniel and Ho, Jonathan and Tarlow, Daniel and Shayer, Peter and Parmar, Niki and Isola, Phillip and Thomas, Danilo J},
  booktitle={Advances in Neural Information Processing Systems},
  volume={34},
  pages={17984--17997},
  year={2021}
}

@inproceedings{singh2023retrievaldiffusion,
author = {Blattmann, Andreas and Rombach, Robin and Oktay, Kaan and M\"{u}ller, Jonas and Ommer, Bj\"{o}rn},
title = {Retrieval-augmented diffusion models},
year = {2022},
isbn = {9781713871088},
publisher = {Curran Associates Inc.},
address = {Red Hook, NY, USA},
booktitle = {Proceedings of the 36th International Conference on Neural Information Processing Systems},
articleno = {1114},
numpages = {16},
location = {New Orleans, LA, USA},
series = {NIPS '22}
}

@inproceedings{
sahoo2024mdlm,
title={Simple and Effective Masked Diffusion Language Models},
author={Subham Sekhar Sahoo and Marianne Arriola and Aaron Gokaslan and Edgar Mariano Marroquin and Alexander M Rush and Yair Schiff and Justin T Chiu and Volodymyr Kuleshov},
booktitle={The Thirty-eighth Annual Conference on Neural Information Processing Systems},
year={2024},
url={https://openreview.net/forum?id=L4uaAR4ArM}
}

@misc{nie2025llada,
      title={Large Language Diffusion Models}, 
      author={Shen Nie and Fengqi Zhu and Zebin You and Xiaolu Zhang and Jingyang Ou and Jun Hu and Jun Zhou and Yankai Lin and Ji-Rong Wen and Chongxuan Li},
      year={2025},
      eprint={2502.09992},
      archivePrefix={arXiv},
      primaryClass={cs.CL},
      url={https://arxiv.org/abs/2502.09992}, 
}

@inproceedings{zeng-etal-2025-bridging,
    title = "Bridging the Editing Gap in {LLM}s: {F}ine{E}dit for Precise and Targeted Text Modifications",
    author = "Zeng, Yiming  and
      Yu, Wanhao  and
      Li, Zexin  and
      Ren, Tao  and
      Ma, Yu  and
      Cao, Jinghan  and
      Chen, Xiyan  and
      Yu, Tingting",
    editor = "Christodoulopoulos, Christos  and
      Chakraborty, Tanmoy  and
      Rose, Carolyn  and
      Peng, Violet",
    booktitle = "Findings of the Association for Computational Linguistics: EMNLP 2025",
    month = nov,
    year = "2025",
    address = "Suzhou, China",
    publisher = "Association for Computational Linguistics",
    url = "https://aclanthology.org/2025.findings-emnlp.118/",
    doi = "10.18653/v1/2025.findings-emnlp.118",
    pages = "2193--2206",
    ISBN = "979-8-89176-335-7"
}

@misc{shu2023rewrite,
      title={RewriteLM: An Instruction-Tuned Large Language Model for Text Rewriting}, 
      author={Lei Shu and Liangchen Luo and Jayakumar Hoskere and Yun Zhu and Yinxiao Liu and Simon Tong and Jindong Chen and Lei Meng},
      year={2023},
      eprint={2305.15685},
      archivePrefix={arXiv},
      primaryClass={cs.CL},
      url={https://arxiv.org/abs/2305.15685}, 
}

@misc{zhao2025surveylargelanguagemodels,
      title={A Survey of Large Language Models}, 
      author={Wayne Xin Zhao and Kun Zhou and Junyi Li and Tianyi Tang and Xiaolei Wang and Yupeng Hou and Yingqian Min and Beichen Zhang and Junjie Zhang and Zican Dong and Yifan Du and Chen Yang and Yushuo Chen and Zhipeng Chen and Jinhao Jiang and Ruiyang Ren and Yifan Li and Xinyu Tang and Zikang Liu and Peiyu Liu and Jian-Yun Nie and Ji-Rong Wen},
      year={2025},
      eprint={2303.18223},
      archivePrefix={arXiv},
      primaryClass={cs.CL},
      url={https://arxiv.org/abs/2303.18223}, 
}

@article{arriola2025block,
  title={Block diffusion: Interpolating between autoregressive and diffusion language models},
  author={Arriola, Marianne and Gokaslan, Aaron and Chiu, Justin T and Yang, Zhihan and Qi, Zhixuan and Han, Jiaqi and Sahoo, Subham Sekhar and Kuleshov, Volodymyr},
  journal={arXiv preprint arXiv:2503.09573},
  year={2025}
}

@misc{chen2025codacodinglmdiffusion,
      title={CoDA: Coding LM via Diffusion Adaptation}, 
      author={Haolin Chen and Shiyu Wang and Can Qin and Bo Pang and Zuxin Liu and Jielin Qiu and Jianguo Zhang and Yingbo Zhou and Zeyuan Chen and Ran Xu and Shelby Heinecke and Silvio Savarese and Caiming Xiong and Huan Wang and Weiran Yao},
      year={2025},
      eprint={2510.03270},
      archivePrefix={arXiv},
      primaryClass={cs.LG},
      url={https://arxiv.org/abs/2510.03270}, 
}

@misc{dream2025,
      title={Dream-Coder 7B: An Open Diffusion Language Model for Code}, 
      author={Zhihui Xie and Jiacheng Ye and Lin Zheng and Jiahui Gao and Jingwei Dong and Zirui Wu and Xueliang Zhao and Shansan Gong and Xin Jiang and Zhenguo Li and Lingpeng Kong},
      year={2025},
      eprint={2509.01142},
      archivePrefix={arXiv},
      primaryClass={cs.CL},
      url={https://arxiv.org/abs/2509.01142}, 
}

@inproceedings{evalplus,
  title = {Is Your Code Generated by Chat{GPT} Really Correct? Rigorous Evaluation of Large Language Models for Code Generation},
  author = {Liu, Jiawei and Xia, Chunqiu Steven and Wang, Yuyao and Zhang, Lingming},
  booktitle = {Thirty-seventh Conference on Neural Information Processing Systems},
  year = {2023},
  url = {https://openreview.net/forum?id=1qvx610Cu7}
}

@inproceedings{
wei2022chain,
title={Chain of Thought Prompting Elicits Reasoning in Large Language Models},
author={Jason Wei and Xuezhi Wang and Dale Schuurmans and Maarten Bosma and brian ichter and Fei Xia and Ed H. Chi and Quoc V Le and Denny Zhou},
booktitle={Advances in Neural Information Processing Systems},
editor={Alice H. Oh and Alekh Agarwal and Danielle Belgrave and Kyunghyun Cho},
year={2022},
url={https://openreview.net/forum?id=_VjQlMeSB_J}
}

@inproceedings{
hu2022lora,
title={Lo{RA}: Low-Rank Adaptation of Large Language Models},
author={Edward J Hu and yelong shen and Phillip Wallis and Zeyuan Allen-Zhu and Yuanzhi Li and Shean Wang and Lu Wang and Weizhu Chen},
booktitle={International Conference on Learning Representations},
year={2022},
url={https://openreview.net/forum?id=nZeVKeeFYf9}
}

@article{T5Raffel,
author = {Raffel, Colin and Shazeer, Noam and Roberts, Adam and Lee, Katherine and Narang, Sharan and Matena, Michael and Zhou, Yanqi and Li, Wei and Liu, Peter J.},
title = {Exploring the limits of transfer learning with a unified text-to-text transformer},
year = {2020},
issue_date = {January 2020},
publisher = {JMLR.org},
volume = {21},
number = {1},
issn = {1532-4435},
journal = {J. Mach. Learn. Res.},
month = jan,
articleno = {140},
numpages = {67}
}

@inproceedings{codefusion,
  title={Codefusion: A pre-trained diffusion model for code generation},
  author={Singh, Mukul and Cambronero, Jos{\'e} and Gulwani, Sumit and Le, Vu and Negreanu, Carina and Verbruggen, Gust},
  booktitle={Proceedings of the 2023 Conference on Empirical Methods in Natural Language Processing},
  pages={11697--11708},
  year={2023}
}

@inproceedings{AST5,
author = {Gong, Linyuan and Elhoushi, Mostafa and Cheung, Alvin},
title = {AST-T5: structure-aware pretraining for code generation and understanding},
year = {2024},
publisher = {JMLR.org},
booktitle = {Proceedings of the 41st International Conference on Machine Learning},
articleno = {635},
numpages = {15},
location = {Vienna, Austria},
series = {ICML'24}
}

@misc{gong2025diffucoder,
      title={DiffuCoder: Understanding and Improving Masked Diffusion Models for Code Generation}, 
      author={Shansan Gong and Ruixiang Zhang and Huangjie Zheng and Jiatao Gu and Navdeep Jaitly and Lingpeng Kong and Yizhe Zhang},
      year={2025},
      eprint={2506.20639},
      archivePrefix={arXiv},
      primaryClass={cs.CL},
      url={https://arxiv.org/abs/2506.20639}, 
}

@inproceedings{
zhao2025d,
title={d1: Scaling Reasoning in Diffusion Large Language Models via Reinforcement Learning},
author={Siyan Zhao and Devaansh Gupta and Qinqing Zheng and Aditya Grover},
booktitle={The Thirty-ninth Annual Conference on Neural Information Processing Systems},
year={2025},
url={https://openreview.net/forum?id=7ZVRlBFuEv}
}

@article{guo2025deepseek,
  title={DeepSeek-R1 incentivizes reasoning in LLMs through reinforcement learning},
  author={Guo, Daya and Yang, Dejian and Zhang, Haowei and Song, Junxiao and Wang, Peiyi and Zhu, Qihao and Xu, Runxin and Zhang, Ruoyu and Ma, Shirong and Bi, Xiao and others},
  journal={Nature},
  volume={645},
  number={8081},
  pages={633--638},
  year={2025},
  publisher={Nature Publishing Group UK London}
}

@article{austin2021program,
  title={Program synthesis with large language models},
  author={Austin, Jacob and Odena, Augustus and Nye, Maxwell and Bosma, Maarten and Michalewski, Henryk and Dohan, David and Jiang, Ellen and Cai, Carrie and Terry, Michael and Le, Quoc and others},
  journal={arXiv preprint arXiv:2108.07732},
  year={2021}
}

@misc{chen2021evaluatinglargelanguagemodels,
      title={Evaluating Large Language Models Trained on Code}, 
      author={Mark Chen and Jerry Tworek and Heewoo Jun and Qiming Yuan and Henrique Ponde de Oliveira Pinto and Jared Kaplan and Harri Edwards and Yuri Burda and Nicholas Joseph and Greg Brockman and Alex Ray and Raul Puri and Gretchen Krueger and Michael Petrov and Heidy Khlaaf and Girish Sastry and Pamela Mishkin and Brooke Chan and Scott Gray and Nick Ryder and Mikhail Pavlov and Alethea Power and Lukasz Kaiser and Mohammad Bavarian and Clemens Winter and Philippe Tillet and Felipe Petroski Such and Dave Cummings and Matthias Plappert and Fotios Chantzis and Elizabeth Barnes and Ariel Herbert-Voss and William Hebgen Guss and Alex Nichol and Alex Paino and Nikolas Tezak and Jie Tang and Igor Babuschkin and Suchir Balaji and Shantanu Jain and William Saunders and Christopher Hesse and Andrew N. Carr and Jan Leike and Josh Achiam and Vedant Misra and Evan Morikawa and Alec Radford and Matthew Knight and Miles Brundage and Mira Murati and Katie Mayer and Peter Welinder and Bob McGrew and Dario Amodei and Sam McCandlish and Ilya Sutskever and Wojciech Zaremba},
      year={2021},
      eprint={2107.03374},
      archivePrefix={arXiv},
      primaryClass={cs.LG},
      url={https://arxiv.org/abs/2107.03374}, 
}
\clearpage
\newpage

\appendix

\section{Implementation Details and Experimental Setup}
\label{sec:appendix_implementation}

In this section, we provide a comprehensive delineation of our experimental framework, focusing on the reproducibility of our results and the specific mechanics of the two distinct masking strategies we investigated. All computational experiments were executed using the PyTorch framework, accelerated by DeepSpeed ZeRO-2 optimization on 8 NVIDIA A100 GPUs. We selected \textit{LLaDA-8B-Instruct} as our backbone model due to its inherent capabilities in sequence modeling, and we fine-tuned it using a consistent set of hyperparameters to ensure a rigorous comparison between our proposed methods.

\subsection{General Training and LoRA Configuration}

To maintain experimental consistency, we utilized a unified optimization configuration across all masking strategies. The model was trained for 3 epochs using the AdamW optimizer with $\beta_1=0.9$ and $\beta_2=0.999$. We employed a cosine learning rate scheduler that includes a 10\% warmup period, allowing the model to stabilize its gradients before reaching the peak learning rate of $5\times 10^{-5}$. To maximize computational throughput while maintaining numerical stability, we conducted all training in \texttt{bfloat16} precision with a maximum sequence length of 4096 tokens. A per-device batch size of 1 was used in conjunction with 16 gradient accumulation steps to simulate a larger effective batch size, ensuring stable convergence.

We applied Low-Rank Adaptation (LoRA) extensively across the model architecture ~\cite{hu2022lora}. Unlike conservative approaches that only target attention projection matrices, we attached LoRA adapters to all linear layers within the transformer blocks, including the query, key, value, and output projections in the self-attention mechanism, as well as the gate, up, and down projections in the MLP feed-forward networks. We set the LoRA rank $r$ to 64 and the scaling factor $\alpha$ to 128, resulting in a robust adaptation capacity. A dropout rate of 0.1 was applied to the LoRA layers to prevent overfitting. 

\paragraph{Inference Setting.}  
We adopt a semi-autoregressive decoding scheme at inference time as illustrated in ~\cite{nie2025llada}.
Given an input prompt, the target sequence is generated in fixed-length blocks, each of which is initialized as fully masked and iteratively refined over multiple diffusion steps.
Within each block, the model predicts all masked positions in parallel and progressively commits a subset of tokens based on confidence, while lower-confidence positions remain masked for further refinement.
Blocks are generated sequentially from left to right to preserve causal structure, resulting in a decoding process that is non-autoregressive within blocks but autoregressive across blocks.
Unless otherwise specified, inference is performed with deterministic decoding (temperature set to zero).

\subsection{Hyperparameter Settings}
\label{hyper_settings}

We utilize a dynamic, curriculum-based approach to all masking method. This strategy is designed to function as a denoising autoencoder objective that evolves in difficulty throughout the training process. Instead of employing a static masking probability, which is common in standard Masked Language Modeling (MLM), we implemented a time-dependent noise schedule. We define an adaptive base noise rate $\epsilon_t$ for each training step $t$, which follows a cosine annealing curve decaying from a maximum noise level $\epsilon_{max}=0.1$ down to a minimum $\epsilon_{min}=0.001$. This schedule ensures that the model is exposed to high-noise regularization during the early phases of training to learn robust feature representations, while gradually transitioning to a low-noise refinement phase towards the end of training to perfect its generation capabilities.

Furthermore, we introduced a stochastic element to the masking intensity at the sample level. For any given step with base rate $\epsilon_t$, the actual masking probability $P_{mask}$ for a specific data sample is not fixed at $\epsilon_t$ but is sampled from a uniform distribution range $[\epsilon_t, 1.0]$. Mathematically, this is implemented as $P_{mask} = (1 - \epsilon_t) \cdot \lambda + \epsilon_t$, where $\lambda \sim U(0, 1)$. This mechanism creates a diverse training environment where, within the same batch, the model may encounter samples that are almost entirely masked alongside samples that retain most of their original tokens. This high variance in partial observability forces the model to rely heavily on long-range context and internal reasoning chains to reconstruct the missing information, rather than relying on local surface-level cues.

To ensure the noise distribution reflects the structural importance of various code elements, we employ a category-aware weighting scheme during the sampling process. Specifically, we define a baseline weight $p_{\text{skel}} = 0.15$ for skeletal tokens and auxiliary syntax such as punctuation and constants. For data-driven components including assignments and function calls, we assign a moderate weight of $p_{\text{data}} = 0.42$. To compel the model to prioritize the reconstruction of the program's execution backbone, we significantly elevate the weights for structural nodes: conditional statements (e.g., \texttt{if}, \texttt{elif}) are set to $p_{\text{cond}} = 0.58$, and the most critical control flow elements (e.g., \texttt{for}, \texttt{while}, \texttt{return}) receive the highest priority with $p_{\text{ctrl}} = 0.60$. These assigned weights for each category $\mathcal{P} = \{p_{\text{skel}}, p_{\text{data}}, p_{\text{cond}}, p_{\text{ctrl}}\}$ are summarized in Table \ref{tab:node_weights}.

\begin{table}[h]
\centering
\small
\begin{tabularx}{\columnwidth}{Xcc} 
\toprule
\textbf{Category} & \textbf{Symbol} & \textbf{Weight} \\ \midrule
Skeletal tokens   & $p_{\text{skel}}$ & 0.15 \\
Data logic        & $p_{\text{data}}$ & 0.42 \\
Conditionals      & $p_{\text{cond}}$ & 0.58 \\
Control flow      & $p_{\text{ctrl}}$ & 0.60 \\ \bottomrule
\end{tabularx}
\caption{Sampling Weights for Node Categories.}
\label{tab:node_weights}
\end{table}

\subsection{Dataset Processing Pipeline}

Both methods were evaluated using the \texttt{nvidia/OpenCodeReasoning} dataset. We developed a specialized data processing pipeline to accommodate the structural requirements of our experiments. For each raw sample, we constructed a prompt using a standard chat template and explicitly parsed the assistant's response to isolate the Chain-of-Thought (CoT) reasoning block (delimited by \texttt{<think>} tags) from the executable code solution. This segmentation was critical for our experiments, particularly for Method II, as it allowed us to apply the AST parser exclusively to the valid Python code regions without attempting to parse natural language reasoning steps, which would result in syntax errors. The prompt length was dynamically calculated to ensure that user instructions were never masked, focusing the loss computation solely on the model's generative output.

\section{Qualitative Study}
\begin{algorithm}[h]
\caption{AST-Guided Masking with Expected Span Corruption}
\label{alg:ast_span_v2}
\begin{algorithmic}[1]
\Require $x_{0} \in \mathcal{V}^{L}$ \Comment{token sequence}
\Require $\mathcal{S}_{x} = \{(s_i, e_i, t_i)\}_{i=1}^{n}$ \Comment{AST spans with Types $t_i$}
\Require $\epsilon_{t} \in [0, 1]$ \Comment{target corruption rate}
\Require $\mathcal{W}_{\text{tier}}$ \Comment{Tier weights from Alg. 1}
\Ensure $x_{t}$

\State $N \leftarrow \lfloor \epsilon_{t} \cdot L \rfloor$ \Comment{target \#masked tokens}
\State $m \leftarrow 0^{L}$ \Comment{mask vector}
\State $\mathcal{S}_{cand} \leftarrow \text{Shuffle}(\mathcal{S}_{x})$ \Comment{Randomize traverse order}
\State $c \leftarrow 0$ \Comment{current masked-token count}

\State // Phase 1: span-level masking on code regions
\For{$(s, e, type) \in \mathcal{S}_{cand}$}
    \State $l \leftarrow e - s$
    \State $p \leftarrow \mathcal{W}_{\text{tier}}[type]$ \Comment{Calculate $p$ by Tier Weight} \label{line:calc_p}
    \If{$\text{Bernoulli}(p) = 1$ \textbf{and} $m[s:e]$ has no 1}
        \State $m[s:e] \leftarrow 1$
        \State $c \leftarrow c + l$
        \If{$c \ge N$}
            \State \textbf{break}
        \EndIf
    \EndIf
\EndFor

\State // Phase 2: fallback token masking (Same as Alg. 1)
\If{$c < N$}
    \State $\mathcal{I}_{\text{unmasked}} \leftarrow \{i \mid m[i] = 0\}$
    \State Randomly pick $N - c$ indices from $\mathcal{I}_{\text{unmasked}}$ set to 1
\EndIf

\State // Phase 3: materialize
\State $x_{t} \leftarrow \langle \text{mask} \rangle \odot m + x_{0} \odot (1 - m)$
\State \Return $x_{t}$
\end{algorithmic}
\end{algorithm}
\paragraph{Comparison: HumanEval/54}
Given two strings s0 and s1,
return True if they contain the same set of characters, and False otherwise.

\vspace{0.8em}

\begin{lstlisting}[language=Python,basicstyle=\small]
# AST
set0 = set(s0)
set1 = set(s1)
return set0 == set1
\end{lstlisting}

\begin{lstlisting}[language=Python,basicstyle=\small]
# random masking
return sorted(s0) == sorted(s1)
\end{lstlisting}

\vspace{0.8em}

In {HumanEval/54}, the goal is to check whether two strings contain the same set of characters, regardless of frequency. The AST-aware model correctly applies \texttt{set()} conversion and equality comparison, directly aligning with the task semantics. The random masking model, however, uses \texttt{sorted()} comparison, which implicitly assumes that character multiplicities are identical. As a result, it performs an anagram check rather than a true set identity comparison. This reflects a tendency toward pattern overgeneralization, often seen in token-level models trained without structural guidance. In contrast, the AST-trained model captures the problem intent more faithfully by reconstructing logic at the level of semantically meaningful code blocks.

\paragraph{Comparison: HumanEval/74}
You are given two lists of strings lst1 and lst2.
Return the list that has a shorter total number of characters.
If both have the same total character count, return lst1.
\vspace{0.8em}

\begin{lstlisting}[language=Python,basicstyle=\small]
# AST
total_chars1 = sum(len(s) for s in lst1)
total_chars2 = sum(len(s) for s in lst2)
if total_chars1 < total_chars2:
    return lst1
elif total_chars1 > total_chars2:
    return lst2
else:
    return lst1
\end{lstlisting}

\begin{lstlisting}[language=Python,basicstyle=\small]
# random masking
if len(lst1) < len(lst2):
    return lst1
elif len(lst1) > len(lst2):
    return lst2
else:
    total1 = sum(len(s) for s in lst1)
    total2 = sum(len(s) for s in lst2)
    if total1 < total2:
        return lst1
    elif total1 > total2:
        return lst2
    else:
        return lst1
\end{lstlisting}

\vspace{0.8em}

In HumanEval/74, the AST-aware model first isolates each list in its own block, iterates through each string to sum characters, then compares totals and handles ties in scoped steps. Operations (aggregation, comparison, tie-breaking) reside in dedicated segments, keeping logic transparent and compositional. In contrast, random-masking baseline shortcuts by comparing the lengths of the lists, a poor proxy of character count, then backtracks to calculate string lengths, introducing redundant branches and obscuring strategy. This inconsistency stems from its lack of structural bias. With AST-derived priors, the AST-aware model decomposes tasks into modular, semantically aligned operations, yielding robust, interpretable reasoning with clear logic boundaries.

\paragraph{Comparison: HumanEval/133}
You are given a list of numbers.
You need to return the sum of squared numbers in the given list,
round each element in the list to the upper int (ceiling) first.

\vspace{0.3em}

\begin{lstlisting}[language=Python,basicstyle=\small]
# AST
import math
rounded_lst = [math.ceil(x) for x in lst]
squared_lst = [x**2 for x in rounded_lst]
return sum(squared_lst)
\end{lstlisting}
\begin{lstlisting}[language=Python,basicstyle=\small]
# random masking
rounded_lst = [math.ceil(x) for x in lst]
squared_lst = [x**2 for x in rounded_lst]
return sum(squared_lst)
\end{lstlisting} 

\begin{figure}
    \centering
    \includegraphics[width=1.0\linewidth]{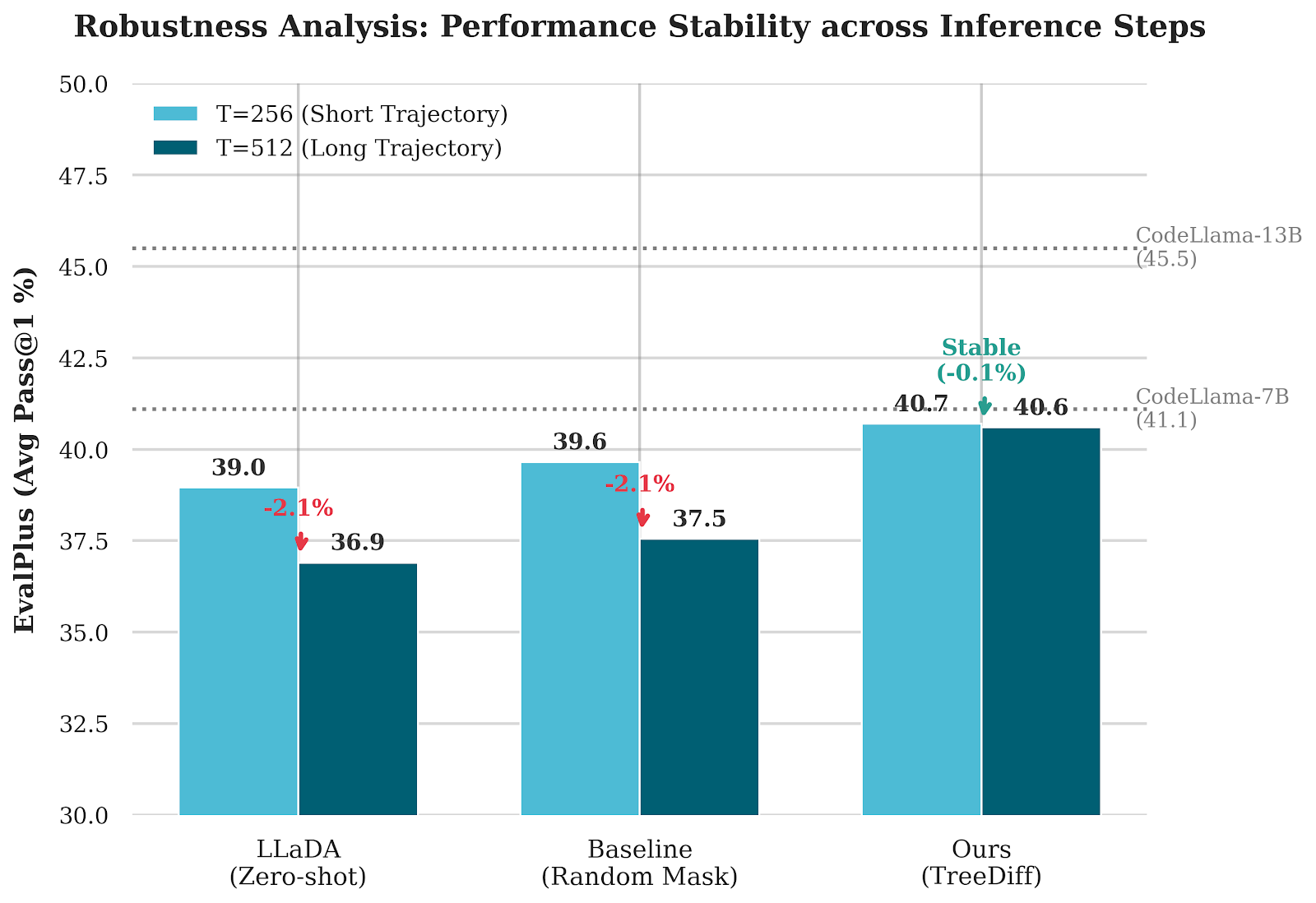}
    \caption{Robustness Analysis}
    \label{robustness}
\end{figure}

In HumanEval/133, the task involves rounding each number in a list to the ceiling, squaring them, and summing the results. The AST‐aware model generates a truly modular and executable solution: it explicitly imports the math module, applies math.ceil elementwise (even handling empty lists gracefully), then computes the sum of squares—each step encapsulated in its own, semantically coherent block. This reflects the model’s structural generalization ability learned via AST‐guided corruption, yielding readable, maintainable code. In contrast, the random‐masking model omits the import and inlines operations without clear boundaries, suggesting reliance on partially memorized token patterns rather than grounded reconstruction. Its output may look correct in isolation, but would fail at runtime without implicit context, highlighting why training models to recover full program structure, not just surface token spans, is essential for robust code generation.
\paragraph{Robustness Analysis}
In addition to the main results, we evaluate the robustness of \texttt{TreeDiff} by varying the number of inference steps $T$ during the denoising process. Figure \ref{robustness} compares the performance of \texttt{TreeDiff} against the \texttt{Baseline (Random Mask)} and \texttt{LLaDA (Zero-shot)} at $T=256$ and $T=512$.As illustrated, standard diffusion-based code generators exhibit a noticeable performance degradation when the inference trajectory is extended. Specifically, both the \texttt{Baseline} and \texttt{LLaDA} experience a 2.1\% absolute drop in the EvalPlus metric as $T$ increases from 256 to 512, indicating that longer denoising chains without structural guidance often lead to accumulated syntactic errors or logical drift.In contrast, \texttt{TreeDiff} maintains a remarkably stable performance, with a negligible decrease of only 0.1\% (from 40.7\% to 40.6\%). This stability underscores the effectiveness of our AST-guided masking strategy. By anchoring the denoising process with crucial structural nodes (e.g., control flows and function skeletons), \texttt{TreeDiff} preserves the program's logical integrity even across longer sampling trajectories. Notably, \texttt{TreeDiff} consistently outperforms the baselines and significantly narrows the gap with the autoregressive \texttt{CodeLlama-7B} model.
\section{AST span-level Masking Strategy}
 As Algorithm \ref{alg:ast_span_v2} shows, let x0 denote the input sequence of tokens; for each AST node, we extract a span (si, ei) indicating the start and end
positions (inclusive) of the corresponding code fragment in
x0. This corresponds to Lines 5–17 in Algorithm 1, where
each span is iterated and its masking probability computed.
These spans provide semantically coherent regions of code
that we later use for structured corruption. Compared to random masking, AST-aware spans preserve the syntactic integrity of code fragments, enabling the model to focus on
reconstructing meaningful program units.

\end{document}